\documentclass[journal]{IEEEtran}

\ifCLASSINFOpdf
   \usepackage[pdftex]{graphicx}
   \graphicspath{{../ICRA2012/figs/}{../jpeg/}}
\else
\fi
\usepackage[cmex10]{amsmath}
\usepackage{caption}
\usepackage{xspace}
\usepackage{algorithm}
\usepackage{algorithmic}
\usepackage{graphicx}
\usepackage{url}
\usepackage{multirow}
\usepackage{wrapfig}
\usepackage{subfig}

\usepackage{amsmath}
\usepackage{amssymb}

\usepackage[margins]{trackchanges} 

\newsubfloat[position=top, listofformat=subsimple]{algorithm}

\newcommand{\secspacea}{\vspace{-0mm}\xspace}
\newcommand{\secspaceb}{\vspace{-0mm}\xspace}
\newcommand{\subsecspacea}{\vspace{-0mm}\xspace}
\newcommand{\subsecspaceb}{\vspace{-0mm}\xspace}

\addeditor{OS} \addeditor{BR} \addeditor{MH} \addeditor{DH} 

\newcommand{\boost}{\textsc{Boost}\xspace}
\newcommand{\cgal}{\textsc{Cgal}\xspace}

\newcommand{\CC}{C\raise.08ex\hbox{\tt ++}\xspace}

\newcommand{\CS}{C-space\xspace}
\newcommand{\CSs}{C-spaces\xspace}
\newcommand{\Cfree}{\ensuremath{\calC_{\rm free}}\xspace}
\newcommand{\Cforb}{\ensuremath{\calC_{\rm forb}}\xspace}
\newcommand{\MMS}{MMS\xspace}
\newcommand{\FSC}{FSC\xspace}
\newcommand{\FSCs}{FSCs\xspace}
\newcommand{\DOF}{\textit{dof}\xspace}
\newcommand{\DOFs}{\textit{dofs}\xspace}

\newcommand{\R}{\ensuremath{\mathbb{R}}}

\newcommand{\PR}{\ensuremath{\mathbb{P(R)}}}

\newcommand{\ie}{{i.e.}\xspace}

\newcommand{\calC}{\ensuremath{\mathcal{C}}\xspace}

\newcommand{\calG}{\ensuremath{\mathcal{G}}\xspace}

\newcommand{\calM}{\ensuremath{\mathcal{M}}\xspace}

\newcommand{\calW}{\ensuremath{\mathcal{W}}\xspace}

\newcommand{\calH}{\ensuremath{\mathcal{H}}\xspace}

  \renewenvironment{thebibliography}[1]{%
    \begin{oldthebibliography}{#1}%
      \footnotesize
      \setlength{\parskip}{0ex}%
      \setlength{\itemsep}{0ex}%
  }%
  {%
    \end{oldthebibliography}%
  }

\newcommand{\ignore}[1]{}
\newcommand{\first}[2]{#1}
\newcommand{\second}[2]{#2}
\newcommand{\journal}[2]{#2} 

\def\marrow{\marginpar[\hfill$\longrightarrow$]{$\longleftarrow$}}

\first{
\def\dan#1{\textcolor{red}{\textsc{Danny says: }{\marrow\sf #1}}}
\def\mike#1{\textcolor{blue}{\textsc{Mike says: }{\marrow\sf #1}}}
\def\barak#1{\textcolor{cyan}{\textsc{Barak says: }{\marrow\sf #1}}}
\def\oren#1{\textcolor{magenta}{\textsc{Oren says: }{\marrow\sf #1}}}
}{
\newcommand{\dan}[1]{}
\newcommand{\mike}[1]{}
\newcommand{\barak}[1]{}
\newcommand{\oren}[1]{}
}

\usepackage{amsthm}
\newtheorem{theorem}{Theorem}[section]
\newtheorem{lemma}[theorem]{Lemma}

\newtheorem{definition}[theorem]{Definition}

\begin{document}

\second{
%
\title{Bare Demo of IEEEtran.cls for Journals}
%
%
%

\author{Michael~Shell,~\IEEEmembership{Member,~IEEE,}
        John~Doe,~\IEEEmembership{Fellow,~OSA,}
        and~Jane~Doe,~\IEEEmembership{Life~Fellow,~IEEE}
\thanks{M. Shell is with the Department
of Electrical and Computer Engineering, Georgia Institute of Technology, Atlanta,
GA, 30332 USA e-mail: (see http://www.michaelshell.org/contact.html).}
\thanks{J. Doe and J. Doe are with Anonymous University.}
\thanks{Manuscript received April 19, 2005; revised January 11, 2007.}}
%
%
}{
\title{On the Power of Manifold Samples in Exploring Configuration Spaces \\
 and the Dimensionality of Narrow Passages}

\newcommand{\email}[1]{{\tt #1}\xspace}

\author{Oren Salzman, Michael Hemmer, and Dan Halperin
\thanks{This work has been supported in part by the 7th Framework
Programme for Research of the European Commission, under
FET-Open grant number 255827 (CGL---Computational Geometry
Learning), by the Israel Science Foundation (grant no.
1102/11), and by the Hermann Minkowski--Minerva Center for
Geometry at Tel Aviv University.}
\thanks{%
Oren Salzman is with the School for Computer Science, 
Tel-Aviv University, Tel Aviv 69978, ISRAEL.
\email{orenzalz@post.tau.ac.il}}%
\thanks{
Michael Hemmer  is with the Institute of Operating Systems and Computer Networks, 
University of Technology Braunschweig, 38106 Braunschweig, GERMANY.
\email{mhsaar@googlemail.com}}%
\thanks{
Dan Halperin is with the School for Computer Science, 
Tel-Aviv University, Tel Aviv 69978, ISRAEL.
\email{danha@post.tau.ac.il}}%
}

}

\markboth{Journal of \LaTeX\ Class Files,~Vol.~6, No.~1, January~2007}%
{Shell \MakeLowercase{\textit{et al.}}: Bare Demo of IEEEtran.cls for Journals}
%


\maketitle

\begin{abstract}
We extend our study of Motion Planning via Manifold Samples (\MMS), 
a general algorithmic framework that combines geometric methods for 
the exact and complete analysis of low-dimensional configuration spaces 
with sampling-based approaches that are appropriate for higher dimensions.  
The framework explores the configuration space by taking samples that are 
\emph{low-dimensional manifolds of the configuration space}
capturing its connectivity much 
better than isolated point samples.
The scheme is particularly suitable for applications in manufacturing, such as assembly planning, where typically motion planning needs to be carried out in very tight quarters.
The contributions of this paper are as follows: 
(i)~We present a recursive application of MMS in a six-dimensional
configuration space, enabling the coordination of two polygonal 
robots translating and rotating amidst polygonal obstacles.
In the adduced experiments for the more demanding test cases 
MMS clearly outperforms PRM, with over 40-fold speedup in a six-dimensional
\emph{coordination-tight} setting. 
(ii)~A probabilistic completeness proof for the case of MMS with samples that are affine subspaces. 
(iii)~A closer examination of the test cases reveals that MMS has,
in comparison to standard sampling-based algorithms,
a significant advantage in scenarios containing 
\emph{high-dimensional narrow passages}. 
This provokes a novel characterization of narrow passages, which 
attempts to capture their dimensionality, an attribute that 
had been (to a large extent) unattended in previous definitions.
\end{abstract}
\textbf{
Note to practitioners---Highly constrained motion-planning scenarios, 
even of low degree of freedom, arise in various applications such as 
assembly planning and manufacturing applications.
Our approach, which emphasizes high precision over any known 
sampling-based technique that we are aware of, allows to cope 
with exactly such cases.
For instance, we show that our framework can be applied
to tight scenarios that arise in three-handed assembly planning.
The ability to cope with tight scenarios is possible, in part, due to recent 
improvements in exact geometric software such as the publicly available 
Computational Geometry Algorithms Library~\cite{cgal} (CGAL).
}



%
\IEEEpeerreviewmaketitle

\section{Introduction}
\secspaceb
\label{sec:introduction}

Configuration spaces, or \CSs, are fundamental tools for studying a large 
variety of systems.
A point in a $d$-dimensional \CS describes one state (or configuration) of a system governed by $d$ parameters.
\CSs appear in diverse domains such as graphical animation, 
surgical planning, computational biology and computer games. 
For a general overview of the subject and its applications 
see, e.g.,~\cite{planning-survey-choset,Latombe-Robot-Mot-Plan, planning-survey-lavalle}. 
The most typical and prevalent example are \CSs describing mobile systems 
(``robots'') with $d$ degrees of freedom (\DOF{s}) moving in some 
\emph{workspace} amongst obstacles. 
As every point in the configuration space \calC corresponds to a \emph{free} or \emph{forbidden} 
pose of the robot, \calC decomposes into disjoint sets $\Cfree$ and $\Cforb$, respectively.
Thus, the  motion-planning problem is commonly reduced to the problem of 
finding a path that is fully contained within $\Cfree$. 

\subsecspacea
\subsection{Background}
\subsecspaceb
\CSs for motion planning haven been intensively studied for over three decades.
Fundamentally, two major approaches exist:

\noindent
\textbf{(i)~Analytic solutions:}
The theoretical foundations, such as the introduction 
of \CSs~\cite{LP80} and the understanding that constructing a \CS 
is computationally hard with respect to the 
number of~\DOFs~\cite{Piano-complexity}, were already laid in 
the late 1970's and early 1980's in the context of motion planing. 
Exact analytic solutions to the general motion-planning 
problem as well as for various low-dimensional instances 
have been proposed in~\cite{Basu03,Canny-complexity,Chazelle199177,PianoII} 
and~\cite{AronovS97,AvnaimBF88,HS96,LP80,PianoI}, respectively. 
For a survey of related approaches see~\cite{planning-survey-sharir}.
However, only recent advances in applied aspects of computational geometry
made robust implementations for important building blocks available. 
For instance, Minkowski sums, which allow the representation of the \CS 
of a translating robot, have robust and exact two- and three-dimensional 
implementations~\cite{fogel-mink,Hachenberger09,w-eecpm-06}. 
Likewise, implementations of planar 
arrangements\footnote{A subdivision of the plane into zero-dimensional, 
one-dimensional and two-dimensional cells, called vertices, edges and 
faces, respectively induced by the curves.} 
for curves~\cite[C.30]{cgal}~\cite{fhw12}, could be used as essential components in~\cite{PianoII}.

\noindent
\textbf{(ii)~Sampling-based approaches:}
Sampling-based approaches, such as Probabilistic Roadmaps (PRM)~\cite{Kavraki-PRM}, 
Expansive Space Trees (EST)~\cite{Hsu99} and Rapidly-exploring Random Trees 
(RRT)~\cite{Lavalle-RRT}, as well as their many variants,  
aim to capture the connectivity of $\Cfree$ in a graph data structure, 
via random sampling of configurations. 
For a general survey on the approach see~\cite{planning-survey-choset, planning-survey-lavalle}.
As opposed to analytic solutions these approaches are also applicable 
to problems with a large number of \DOF. 
Importantly, the PRM and RRT algorithms were shown to be probabilistically  
complete~\cite{Kavraki98,Kuffner00,LK02}, that is, they are guaranteed 
to find a valid solution, if one exists. 
However, the required running time for finding such a solution cannot 
be computed for new queries at run-time. 
This is especially problematic as these algorithms suffer from high sensitivity 
to the so-called ``narrow passage'' problem, e.g., where the robot is required 
to move in environments cluttered with obstacles, having low clearance.

Though there are also some hybrid approaches~\cite{CG-alg-app,hybrid-disks,hybrid-mink,YangS06} that incorporate both analytic and sampling-based approaches,
it is apparent that the arsenal of currently available motion-planning algorithms 
lacks a general scheme applicable to high-dimensional problems with 
little or low sensitivity to narrow passages.
In~\cite{SHRH11} we introduced a framework for Motion Planning via Manifold Samples (\MMS), which also constitutes a hybrid approach. 
In a three-dimensional \CS it was capable of achieving  twenty-fold 
(and more) speedup factor in running time compared to the PRM algorithm when 
used for planning paths within narrow passages. 
We believe that the speedup presented in~\cite{SHRH11} does not present 
a mere algorithmic advantage for a specific implemented instance but a 
fundamental advantage of the framework when solving scenarios with narrow passages.
The \MMS framework is not the first to consider lower dimensional manifolds of the \CS. 
Several algorithms attempt to sample in the \CS, and project the sample to lower dimensional manifolds (see, e.g.,~\cite{BSK11, S07}); however these algorithms still sample points.
For cases where some dimensions are presumed to be decoupled, such as multi-robot navigation, one can sample each robot's individual \CS (see, e.g.,~\cite{AK10, WKC12}) though these algorithms are typically not applicable when there is a tight coupling between the robots.

This study continues developing the \MMS framework as a tool 
to overcome the gap mentioned in existing motion-planning algorithms. 
We briefly present the scheme and continue to a preliminary discussion 
on applying \MMS in high-dimensional \CSs, which motivates this paper.

\subsecspacea
\subsection{Motion Planning via Manifold Samples}
\begin{figure*}[t]
  \centering
   \begin{center}
       \includegraphics[width=0.65\textwidth]{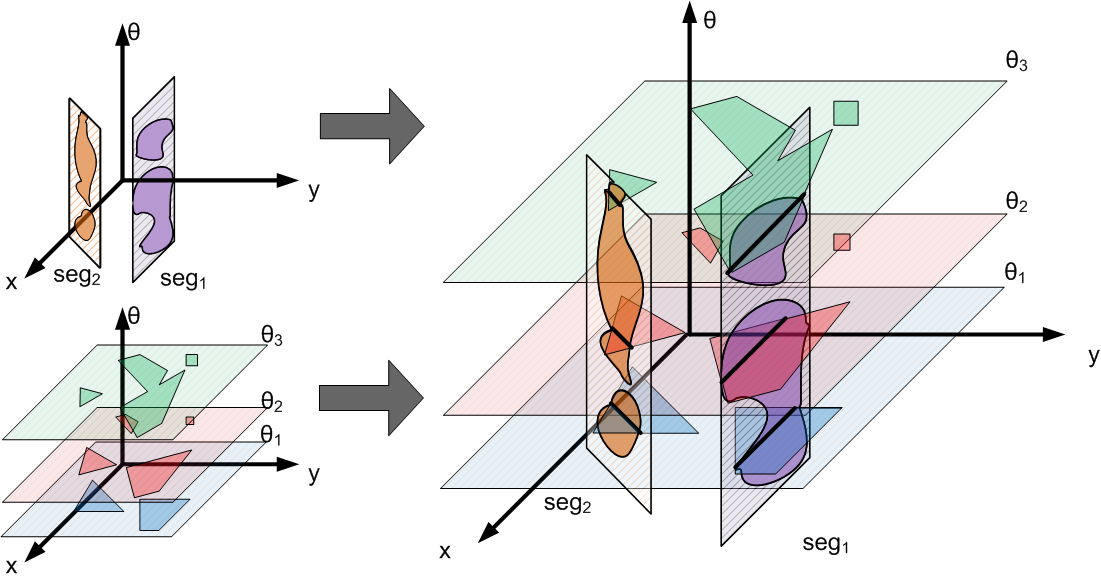}
   \end{center}
   \caption{\sf \MMS in three-dimensional \CSs of translation and rotation in 
   			the plane.
        The left side illustrates two families of manifolds where the 
        decomposed free cells are darkly shaded. The right side illustrates
        their intersection, which induces the graph~$\calG$. 
        Figure taken from~\cite{SHRH11}.}
   \label{fig:conf_space_slicing}
\end{figure*}

\subsecspaceb
The framework is presented as a means to explore the \emph{entire}
\CS, or, in motion-planning terminology as a multi-query planner, 
consisting of a preprocessing stage and a query stage. 
The preprocessing stage constructs the \emph{connectivity graph}~$\calG$ 
of~$\calC$, a data structure that captures the connectivity of~$\calC$ 
using \emph{low-dimensional manifolds} as samples. 
The manifolds are decomposed into cells in~$\Cfree$ and~$\Cforb$ 
in an analytic manner; we call a cell of the decomposed manifold 
that lies in~$\Cfree$ a \emph{free space cell} (\FSC). 
The \FSCs serve as nodes in~$\calG$. Two nodes are connected by an edge 
if their corresponding \FSCs intersect. See Fig.~\ref{fig:conf_space_slicing} 
for an illustration.

Once~$\calG$ has been constructed it can be queried for paths between two  
configurations~$q_s, q_t \in \Cfree$ in the following manner: A manifold that 
contains~$q_s$ in one of its \FSCs is generated and decomposed (similarly for~$q_t$). 
These \FSCs and their appropriate edges are added to~$\calG$.
We compute a path~$\gamma$ (of \FSCs) in~$\calG$ between the \FSCs that contain~$q_s$ and~$q_t$. 
If such a path is found in~$\calG$, it can be (rather straightforwardly) transformed
 into a continuous path in \Cfree by planning a path within each \FSC in~$\gamma$.

\subsecspacea
\subsection{MMS in Higher Dimensions}
\label{subsec:preliminary_discussion}
\subsecspaceb

The successful application of MMS in~\cite{SHRH11} to a three-dimensional \CS 
can be misleading when we come to apply it to higher dimensions.
The heart of the scheme is the choice of manifolds from which we sample. 
Informally, for the scheme to work we must require that the used set of manifolds 
$\calM$ fulfills the following conditions.
\begin{itemize}
\item[\bf C1] The manifolds in $\calM$ cover the \CS.
\item[\bf C2] A pair of surfaces chosen uniformly and 
independently\footnote{The requirement that the choices are independent stems
from the way we prove completeness of the method.
It is not necessarily an essential component of the method itself.}
at random from $\calM$ intersect with significant probability.
\item[\bf C3] 
Manifolds need to be of very low dimension as MMS requires an 
analytic description of the \CS when restricted to a manifold. 
Otherwise the machinery for the
construction of this description is not readily available.
\end{itemize}

For \MMS\ to work in \CSs of dimension $d$, Condition~{\textbf{C2}} 
has a prerequisite that the sum of dimensions of a pair of manifolds 
chosen uniformly and independently at random from $\calM$ is at least $d$ with
significant probability. 
This means in particular that $\calM$ will consist of manifolds of dimension\footnote{The precise statement is somewhat more 
involved and does not contribute much to the informal discussion here. 
Roughly, $\calM$ should comprise manifolds of dimension $\lceil 
\frac{d}{2} \rceil$ or higher and possibly manifolds of their co-dimension.} 
$\lceil \frac{d}{2} \rceil$.
With this prerequisite in mind, there is already much to gain from 
using our existing and strong machinery for analyzing two-dimensional
manifolds~\cite{bfhks-apsca-10, bfhmw-apsgf-10, fhw12}, 
while fulfilling the conditions above: We can solve motion-planning problems 
with four degrees of freedom, at the strength level that \MMS\ offers,
 which is higher than that of standard sampling-based tools. 

However, we wish to advance to higher-dimensional \CSs in which 
satisfying all the above conditions at once is in general impossible.
We next discuss two possible relaxations of the conditions above 
that can lead to effective extensions of \MMS\ to higher dimensions.




\textbf{Dependent choice of manifolds:}
If we insist on using only very low-dimensional manifolds
even in higher-dimensional \CSs, then in order to guarantee that
pairs of manifolds intersect, we need to impose some
dependence between the choices of manifolds, i.e., relaxing condition {\bf C2}.
 A natural way
to impose intersections between manifolds is to adapt the
framework of tree-based planners like RRT~\cite{Lavalle-RRT}. When we
add a new manifold, we insist that it connects either directly or by a sequence
of manifolds to the set of manifolds collected in the data
structure (tree in the case of RRT) so far.

\textbf{Approximating manifolds of high dimension:}
As we do not have the machinery to exactly analyze \CSs
restricted to manifolds of dimension three or higher,
we suggest to substitute exact decomposition of the 
manifolds as induced by the \CS by some approximation.
i.e., relaxing condition {\bf C3}.
There are various ways to carefully approximate C-spaces. 
In the rest of the paper we take the approach of 
{\em a recursive application of \MMS}.


\bigskip


In Section~\ref{sec:two_robots} we demonstrate this recursive 
application for a specific problem in a six-dimensional 
configuration space, namely the coordination of two planar polygonal
robots translating and rotating amidst polygonal obstacles. 
In the adduced experiments for the more demanding test cases 
MMS clearly outperforms several variants and implementations of PRM 
with over 40-fold speedup in an especially tight setting.
Section~\ref{sec:completeness} provides the theoretical 
foundations for using \MMS in a recursive fashion. 
In Section~\ref{sec:discussion} we examine the significant 
advantage of \MMS with respect to prevailing sampling-based approaches 
in scenarios containing \emph{high-dimensional narrow passages}. 
This provokes a novel characterization of narrow passages, which 
attempts to capture their dimensionality.
We conclude with an outlook on further work in Section~\ref{sec:future_work}.

\secspacea
\section{The Case of Two Rigid Polygonal Robots}
\secspaceb
\label{sec:two_robots}

We discuss the \MMS framework applied to the case of coordinating the motion 
of two polygonal robots $R_a$ and $R_b$ translating and rotating in the plane
 amidst polygonal obstacles. 
Each robot is described by the position of its reference point 
$r_a, r_b \in \R^2$
and the amount of counter-clockwise rotation $\theta_a, \theta_b$ 
with respect to an initial orientation. 
All placements of $R_a$ in the workspace~$\calW$ induce the three-dimensional
 space $\calC^a = \R^2 \times S^1$. Similarly for~$R_b$. 
We describe the full system by the six-dimensional \CS 
$\calC = \calC^a \times \calC^b$.

\subsecspacea
\subsection{Recursive Application of the \MMS Framework}
\subsecspaceb
Had we had the means to decompose three-dimensional manifolds the application of \MMS would be straightforward: 
The set $\calM$  consists of two families.
An element of the first family of manifolds is defined by fixing $R_b$ at 
free configurations $b \in \Cfree^b$ while $R_a$ moves freely 
inducing the three-dimensional subspaces\footnote{In this paper, when 
discussing subspaces, unless otherwise stated we refer to affine subspaces 
or  linear manifolds.} 
$\calC^a \times b$. The second family is defined symmetrically by fixing $a\in R_a$.
As subspace pairs of the form 
$(a \times \calC^b, \calC^a \times b)$ 
intersect at the point $(a, b)$, manifolds of the two families 
intersect allowing for connections in the connectivity graph \calG.

\begin{figure*}[t]
  \centering
  \subfloat
   [\sf Horizontal slices]
   {
    \label{fig:angle_primitive}
    \includegraphics[width=0.235\textwidth]{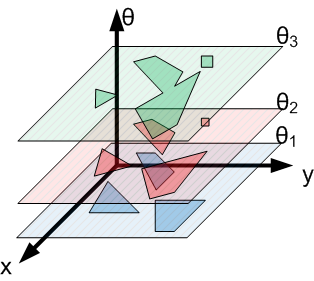}
   }
   \hspace{3mm}
  \subfloat
   [\sf Vertical lines]
   {
    \label{fig:point_primitive}
    \includegraphics[width=0.235\textwidth]{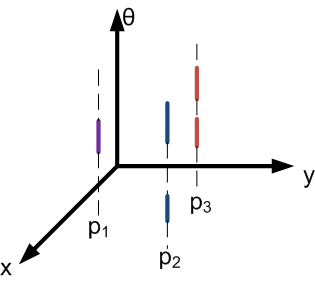}
   }
  \caption{\sf Manifold families and their \FSCs. 
        \FSCs of horizontal slices are polygons while 
        \FSCs of vertical lines are intervals along the line. }
  \label{fig:primitives}
\end{figure*}

However, we do not have the tools to construct three-dimensional manifolds
 explicitly.
Thus the principal idea is to construct \emph{approximations} of these 
manifolds by another application of \MMS.  
Since for a certain manifold one robot is fixed, we are left with a 
three-dimensional \CS in which the fixed robot is regarded as an obstacle. 
Essentially this is done by using the implementation presented in~\cite{SHRH11} 
but with a simpler set of manifolds (see also Fig.~\ref{fig:primitives}):
(i)~{\bf Horizontal slices} -- corresponding to a fixed orientation of the 
moving robot while it is free to translate
(ii)~{\bf Vertical lines} -- corresponding to a fixed location of the reference point of the moving robot while it is free to rotate.

Since we only approximate the three-dimensional subspaces we have to make sure 
that they still intersect. 
In other words, 
if 
$\calC_{\text{apx}}^a$, $\calC_{\text{apx}}^b$ are the approximations of $\calC^a$ and $\calC^b$, respectively, then $(a \times \calC_{\text{apx}}^b, \calC_{\text{apx}}^a \times b)$ 
intersect at the point $(a, b)$ only if $a \in \calC_{\text{apx}}^a$ and $b \in \calC_{\text{apx}}^b$.
To ensure this latter condition we sample an initial set of angles $\Theta_a$ that is used for the first robot 
throughout the entire algorithm.
When approximating its subspace (the second robot is fixed) we take a horizontal slice 
for each angle in $\Theta_a$. At the same time, we only fix the robots position at angles in  $\Theta_a$.
We do the same for the second robot and a set $\Theta_b$. This way it is ensured that 
even the approximations of the three-dimensional subspaces intersect.

\subsecspacea
\subsection{Implementation Details}
\subsecspaceb

\begin{figure*}
  \centering
  \subfloat
   [\sf Random polygons \vspace{0mm}]
   {\label{fig:scenario_1}
   \includegraphics[width=0.25\textwidth]{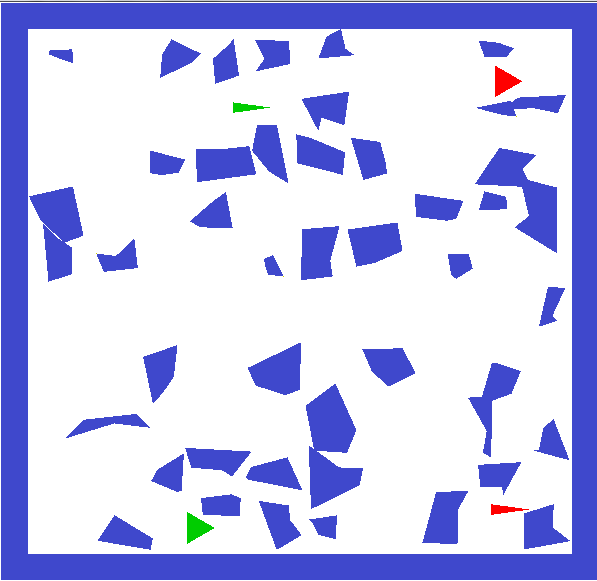}}
  \subfloat
   [\sf Viking helmet  \vspace{0mm}]
   {\label{fig:scenario_2}
   \includegraphics[width=0.25\textwidth]{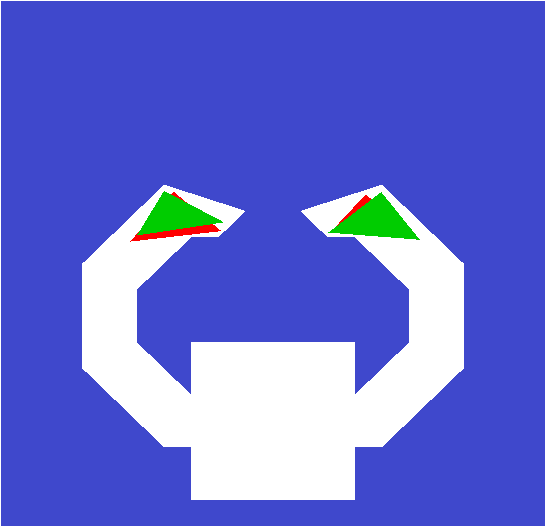}}
  \subfloat      
   [\sf Pacman   \vspace{0mm}]
   {\label{fig:scenario_3}
   \includegraphics[width=0.25\textwidth]{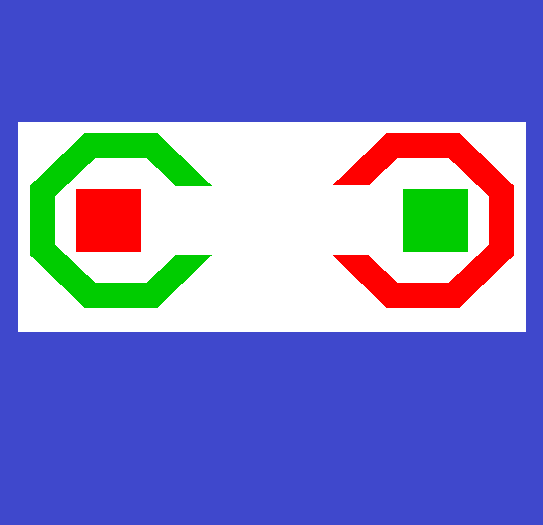}}
  \caption{\sf Experimental scenarios. 
  						 Source and target configurations are drawn in green and red,
  						 respectively.}
  \label{fig:scenarios}
\end{figure*}

\noindent
\textbf{Horizontal slices:}
Let $R_m$ and $R_f$ denote the moving and fixed robot, respectively. 
$\Theta_m$ denotes the set of angles that is sampled for $R_m$. 
A horizontal plane for an angle $\theta_m\in \Theta_m$
is defined by the Minkowski sum of $-R_{\theta_m}$ with all the obstacles and, 
in addition, with the fixed robot.\footnote{$-R_{\theta_m}$ denotes $R_m$ 
rotated around the origin by $\theta_m$ and reflected about the origin.}
However, for each approximation of a three-dimensional affine subspace 
of the robot $R_m$ we are using the same set of angles\footnote{We note that in our implementation, we add a random shift to the set of slices to avoid situations where the initial configuration of one of the robots is aligned with a narrow passage (as is the case in Figure~\ref{fig:scenario_3}).
This is done for each robot independently.}, namely $\Theta_m$.
Only the position of the robot $R_f$ changes.  
Therefore, for all $\theta_m\in \Theta_m$ we precompute the Minkowski sum
of $-R_{\theta_m}$ with all the obstacles. 
In order to obtain a concrete slice we only need to add 
the Minkowski sum of  $-R_{\theta_m}$ with $R_f$. This can be done by a simple overlay operation (see, e.g.,~\cite[C.6]{fhw12}).

\noindent
\textbf{Vertical lines:}
Fixing the reference point of $R^m$ to some location while it is free to rotate induces 
a vertical line in the three-dimensional \CS.
Each vertex (or edge) of the robot in combination
with each edge (or vertex) of an obstacle (or the fixed robot) 
give rise to up to two critical 
angles on this line. These critical values mark a potential transition 
between \Cforb and \Cfree.
Thus a vertical line is constructed by computing these critical angles 
and the \FSCs are maximal free intervals along this line; 
\journal
{for further details see~\cite{SHH12-arxiv}.} 
{(for further details see the Appendix).} 

\ignore{
The Point Primitive is constructed for two sets of points: 
The first, a set of $n_{\ell}$ points that are picked at random for each 
three-dimensional \CS that is computed.
The second set of points consists of $n_{\ell_0}$ points common to all 
three-dimensional \CSs that are computed. For each robot, the $n_{\ell_0}$ 
points are generated and the Point Primitive is constructed disregarding 
the second robot. We term such a construction for a point $p$ a 
\emph{base c-space line} and denote it $S_0(p)$. Hence, in this second case, 
the Point Primitive for a constraining point $p$, a moving robot $R^m$ and 
a fixed robot $R^f$, which we call a \emph{c-space line} is constructed by 
adding the critical angles induced by $R^f$ and $R^m$ located at $p$ to 
the base c-space line $S_0(p)$. }


\subsecspacea
\subsection{Experimental Results}
\subsecspaceb
\label{subsec:experimental_reults}
We demonstrate the performance of our planner using three different scenarios in six-dimensional \CSs. 
All scenarios consist of a workspace, obstacles, two robots and one query 
(source and target configurations). 
Fig.~\ref{fig:scenarios} illustrates the scenarios where the obstacles are
drawn in blue and the source and target configurations are drawn in green
and red, respectively. 
All reported tests were measured on a Lenovo T420 with a 2.8GHz Intel Core i7-2640M CPU processor and 8GB of memory running with a Windows 7 64-bit OS. 
Preprocessing times are the average of 12 runs excluding the minimal and maximal values.
The algorithm is implemented in \CC based on 
\cgal~\cite{cgal}
and the \boost Graph Library~\cite{boost-bgl-2001},
which are used 
for the geometric primitives, 
and the connectivity graph~\calG, respectively.


We chose to compare our planner to the implementation of PRM 
provided by OMPL~\cite{OMPL}. In addition we also
compare with 
Obstacle-Based PRM (OB-PRM)~\cite{AW96} and 
Uniformly distributed Obstacle-Based PRM (U-OB-PRM)~\cite{YTEA12}
(also implemented in OMPL), which were shown to perform better than PRM in many scenarios where narrow passages exist.
We manually optimized the parameters of each planner over a concrete set. 
The parameters used by \MMS are: 
$n_\theta$~--~the number of sampled angles;
$n_\ell$~--~the number of vertical lines;
$n_f$~--~the number of times some robot is fixed to a certain configuration 
while the three-dimensional \CS of the other is computed. 
The parameters used for the PRM algorithms  are: 
$k$~--~the number of neighbors to which each milestone should be connected;
res~--~collision-checking resolution.
U-OB-PRM needs additional parameters, the length $l$ of the line-segments sampled in space and the resolution of samples along this line.
Following the results of~\cite{YTEA12} and after validating these parameters, we used the same collision checking-resolution for the resolution and a line-segment of length equal to 10 times the collision-checking resolution.
We found empirically that in order to obtain the best results from U-OB-PRM, we should add uniform samples to the biased ones.
Thus the variant we used samples half of the time uniformly in space while half of the time uses the scheme suggested in~\cite{YTEA12}.
Table~\ref{tbl:results1} summarized the parameters used by each algorithm, the average running time and the standard deviation (denoted by~$t$ and stdev, respectively).
 
\ignore{
We chose to compare our planner to two variants of PRM that are 
provided by OMPL, the Open Motion Planning Library~\cite{ompl}:
The first is the standard implementation where new samples are sampled 
uniformly from the configuration space (we denote this variant simply as PRM). 
The second, obstacle-based PRM or OBPRM~\cite{AW96}, biases samples near the 
boundary of obstacles.  OBPRM was shown to perform better than PRM in many 
scenarios where narrow passages exist. 
We manually optimized the parameters of each planner over a concrete set. 
The parameters for \MMS are: 
$n_\theta$~--~the number of sampled angles;
$n_\ell$~--~the number of vertical lines;
$n_f$~--~the number of times some robot is fixed to a certain configuration 
while the three-dimensional \CS of the other is computed. The parameters
used for the PRM are:
$k$~--~the number of neighbors to which each milestone should be connected;
res~--~collision checking resolution.
The results are summarized in Table~\ref{tbl:results1}.
}

The Random polygons scenario\footnote{A scenario provided as part of the 
OMPL distribution.} is an easy scenario where little coordination is required. 
Both planners require the same amount of time to solve this case. We see that
 even though our planner uses complex primitives, when using the right 
parameters, it can handle simple cases with no overhead when compared to 
the PRM algorithms.

The Viking-helmet scenario consists of two narrow passages that each robot 
needs to pass through. Moreover, coordination is required for the two robots 
to exchange places in the lower chamber. We see that the running times of the 
\MMS implementation are favorable when compared to the PRM implementations. 
Note that although each robot is required to move along a narrow passage, 
the motion along this passage does not require coordination between the  robots. 

\begin{figure}[b]
  \begin{center}
  	\vspace{-3.5mm}
    \includegraphics[width=0.48\textwidth]{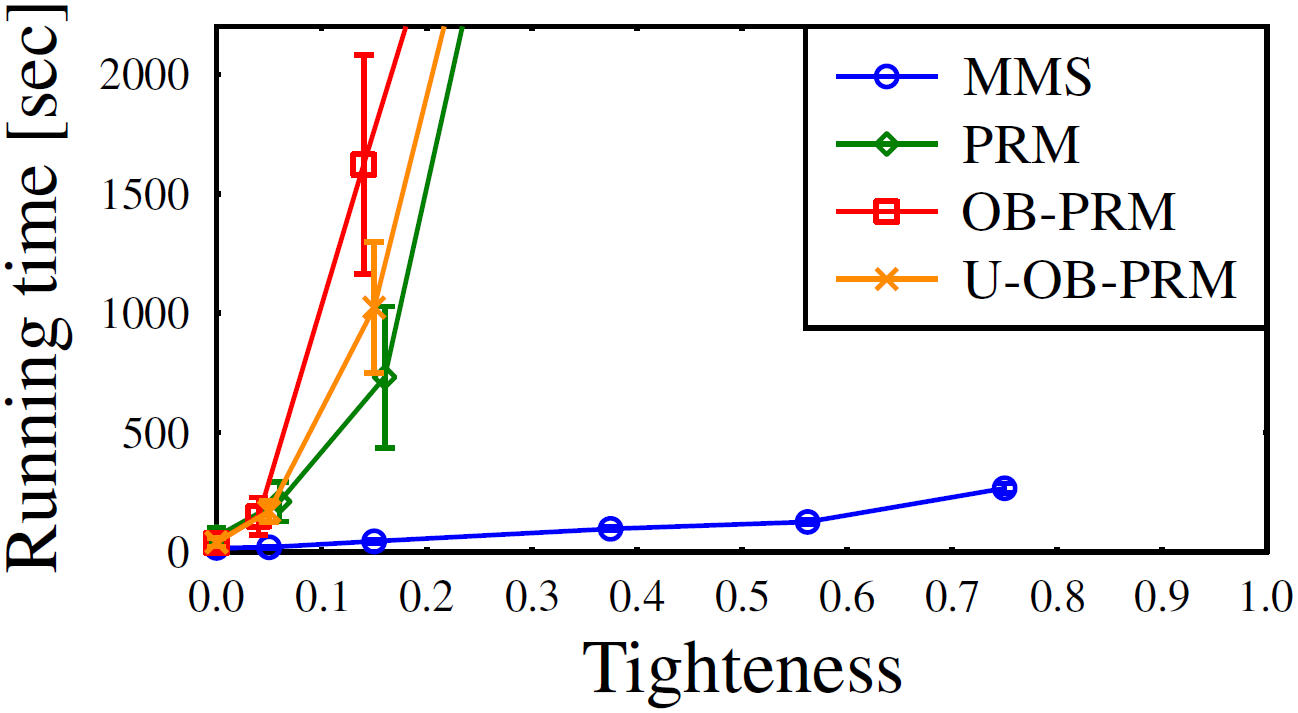}
    \caption{\sf \label{fig:tightness}Tightness Results. 
    							Error bars represent one standard variation.}
  \end{center}
\end{figure}
\begin{figure}
  \centering
  \subfloat
  [\sf The square robot moves into a position where the Pacman can engulf it. \vspace{0mm}]
  {\label{fig:Pacman_motion_1}
    \includegraphics[width=0.3\textwidth]{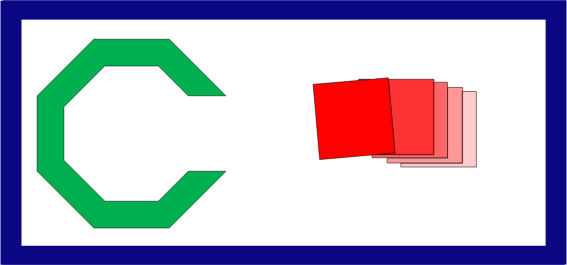}}
  \\
  \subfloat
  [\sf The Pacman engulfs the square robot.\vspace{0mm}]
  {\label{fig:packman_motion_2}
    
    \includegraphics[width=0.3\textwidth]{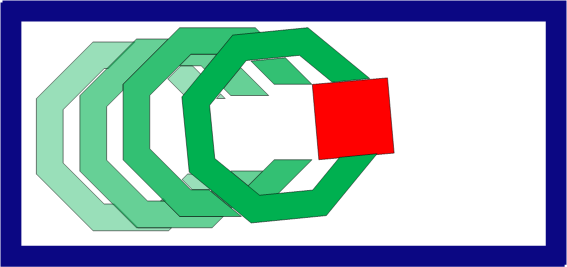}}
  
  
  \caption{\sf Example of a path in the Pacman Scenario.}
  \label{fig:Pacman_motion}
\end{figure}

The Pacman scenario, in which the two robots need to exchange places, 
requires coordination of the robots: they are required to move into a 
position where the C-shaped robot, or Pacman, ``swallows'' the square robot;
the Pacman is then required to rotate around the robot. Finally the two 
robots should move apart (see Fig.~\ref{fig:Pacman_motion}). We ran this 
scenario several times, progressively increasing the square robot size. 
This caused a ``tightening'' of the passages containing the desired path. 
Fig.~\ref{fig:tightness} demonstrates the preprocessing time as a function
 of the tightness of the problem for both planners. A tightness of zero 
denotes the base scenario (Fig.~\ref{fig:scenario_3}) while a tightness 
of one denotes the tightest solvable case. 
Our algorithm is less sensitive to the tightness of the problem when compared to the PRM algorithm. In the tightest experiment solved by all
PRM variants, \MMS runs 10 times faster. We ran the experiment on tighter scenarios 
but all PRM algorithms crashed after 5000 seconds due to lack of memory resources.
We believe that the behavior of the algorithms with respect to the tightness 
of the passage reveals a fundamental difference between the two algorithms and 
discuss this in Section~\ref{sec:discussion}.

%
%

\begin{table*}[tbh]

 \begin{center}
 \begin{tabular}{|c||c|c|c|c|c||c|c|c|c||c|c|c|c||c|c|c|c||}
 \hline
 Scenario   
       & \multicolumn{5}{|c||}{MMS}     
       & \multicolumn{4}{|c||}{PRM} 
       & \multicolumn{4}{|c||}{OB-PRM} 
       & \multicolumn{4}{|c||}{U-OB-PRM}
       \\\cline{2-18}
       &  $n_\theta$ & $n_\ell$ & $n_f$ & t[sec] & stdev
       &  k      & res  & t[sec]  & stdev
       &  k      & res  & t[sec]  & stdev
       &  k      & res  & t[sec]  & stdev
       \\\hline
 Random polygons
       & 5   & 512   & 2  & \textbf{8} 		& 1.6
       & 10  & 0.02  &    	\textbf{14.5} & 8.3
       & 10  & 0.01  &    	\textbf{28.4} & 12
       & 8   & 0.01  &    	\textbf{10.5} & 9.9
       \\
 Viking Helmet
       & 20  & 16    & 10  & 	\textbf{6.2} 	& 1.2
       & 10  & 0.005 &    		\textbf{86.8} & 34
       & 10  & 0.005 &    		\textbf{92.8} & 14
       & 8   & 0.0125&	    	\textbf{40} 	& 28
       \\
 Pacman   
 			 & 5   & 4      & 180  &  \textbf{17.6} & 3.5
       & 12  & 0.015  &    			\textbf{15} 	& 9.5
       & 10  & 0.01   &    			\textbf{18.7} & 6.8
       & 10  & 0.0125 & 				\textbf{20}		& 3.3
       \\\hline
 \end{tabular}
 \end{center}
  \caption{\label{tbl:results1} Comparison of \MMS with PRM variants }

\end{table*}

\secspacea
\section{Probabilistic Completeness of \MMS}
\secspaceb
\label{sec:completeness}
An algorithm is \emph{probabilistically complete} if the probability that it will produce a solution (when one exists) approaches one as more time is spent.
It has been shown that PRM, using point samples, is probabilistically complete 
(see, e.g.,~\cite[C.7]{planning-survey-choset}).
At first glance it may seem that if the scheme is complete for point samples 
then it is evidently complete when these samples are substituted with manifold 
samples: manifolds of dimension one or higher guarantee better coverage of the 
configuration space. 
However, there is a crucial difference between PRM and MMS when it comes to
 connectivity. The completeness proof for  PRM relies, among others,
 on the fact that if the straight line segment in the configuration space 
connecting two nearby samples lies in the free space, then the nodes 
corresponding to these two configurations are connected by an edge in the 
roadmap graph. The connectivity in MMS is attained through intersections
 of manifolds, which may require a chain of subpaths on several distinct 
manifolds to connect two nearby free configurations. This is what makes 
the completeness proof for MMS non trivial and is expressed in 
Lemma~\ref{lem:local_path} below.

We present a probabilistic completeness proof for the \MMS framework for 
the case where the \CS \calC is the d-dimensional Euclidean 
space $\R^d$, while MMS is taking samples from two perpendicular affine 
subspaces, 
the sum of dimensions of which is $d$. 
Assuming that the \CS is Euclidean does not impose a real restriction 
as long as the actual \CS can be embedded in a Euclidean space
(see, e.g.~\cite[Section~3.5, Section 7.1.2]{planning-survey-choset},~\cite[Chapters 4-5]{planning-survey-lavalle} or~\cite{K04}). 

Let $A$ and $B$ denote affine subspaces of \calC and let $k$ and $d-k$ be their dimensions, respectively.
As $\calC$ is decomposed into two perpendicular subspaces, 
a point~$p=(a_1,\dots,a_k,b_1,\dots,b_{d-k}) \in \calC$ may be 
represented as the pair of points $(a,b)$ from subspaces $A$ and $B$. 
Under this assumption, the set of manifolds~$\calM$ 
consists of two families of $k$ and $(d-k)$-dimensional
manifolds~$\calM^A$ and~$\calM^B$. 
Family~$\calM^A$ consists of all manifolds that are defined by fixing
a point $a_0 \in A$ while the remaining $d-k$ parameters
are variable; $\calM^B$ is defined symmetrically.
Two manifolds~$m(a)\in \calM^A$ and~$m(b)\in \calM^B$ always 
intersect in exactly one point, \ie,~$m(a)\cap m(b)=(a,b)\in \calC$.  
Let~$B^{\calC}_r(p) = \{q \in \calC~|~dist(p,q)\leq r \}$ define  
a ball in~$\calC$ of radius~$r$ centered at~$p\in\calC$, 
where~$dist$ denotes the Euclidean metric on~$\calC$. 
Likewise,~$B^{B}_r(b)$ and~$B^{A}_r(a)$ denote $(d-k)$ and $k$-dimensional balls 
in~$B$ and~$A$, respectively.

\begin{definition}[$\rho$-intersecting]
For~$\rho>0$ we call a manifold~$m(a)\in \calM^A$~$\rho$-{\em intersecting}
for a point~$p \in \calC$ if~$m(a)\cap B^{\calC}_{\rho}(p) \neq \emptyset$, \ie, if 
$a \in B^{A}_{\rho}(p_A)$, where~$p_A$ is the projection of~$p$ onto~$A$.  
Similarly for manifolds in~$B$. 
\end{definition}

A feasible path~$\gamma$ is a continuous mapping from the 
interval~$[0,1]$ into~$\Cfree$. The image of a path is defined as 
$Im(\gamma) = \{ \gamma(\alpha) \ | \ \alpha \in [0,1] \}$.
We show that for any collision-free path~$\gamma_{p,q}$ of clearance 
$\rho > 0~$ between two configurations~$p$ and~$q$ the MMS constructs a path from~$p$ to~$q$ such that 
(i)~the path lies on the \FSCs of the sampled manifolds and
(ii)~every point on the path is at distance at most $\rho$ from~$\gamma_{p,q}$,
with a positive probability. 
Moreover, the probability of failing to find such a path by the \MMS algorithm 
decreases exponentially with the number of samples.



\begin{figure*}[t,b]
  \centering
  \subfloat
   []
   {\label{fig:ver_hor_intersection}
   \includegraphics[width=0.29\textwidth]{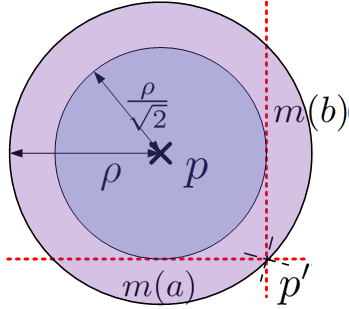}}
   \hspace{10mm}
  \subfloat
   []
   {\label{fig:path_example}
   \includegraphics[width=0.38\textwidth]{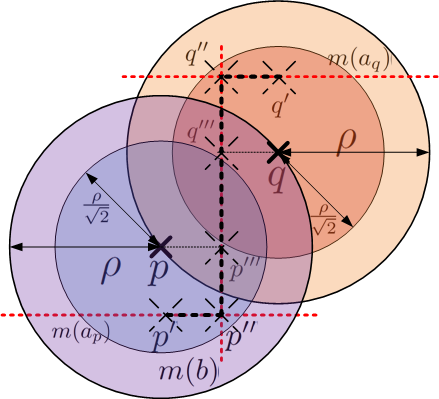}}
  \caption{%
\sf Two-dimensional sketch: balls and manifolds are presented as circles and lines, respectively. 
(a)~Intersection of two $\rho/\sqrt{2}$-intersecting manifolds.  
(b)~Construction of a path as defined in Lemma~\ref{lem:local_path}.
}
\end{figure*}

\begin{lemma}
\label{lem:intersection_point}
For~$p \in \calC$ and~$\rho>0$ let~$m(a)\in \calM^A$ and~$m(b)\in \calM^B$ be two 
manifolds that are~$\rho/\sqrt{2}$-intersecting for~$p$. 
Their intersection point~$p'=(a,b)=m(a)\cap m(b)$ is in~$B^{\calC}_{\rho}(p)$. 
\end{lemma}

\journal
{ 
For an illustration of Lemma~\ref{lem:intersection_point} 
see Fig.~\ref{fig:ver_hor_intersection}.
The proof (given in~\cite{SHH12-arxiv}) follows immediately from elementary 
properties of $\R^d$. 
Lemma~\ref{lem:local_path} is the centerpiece of the completeness proof as it establishes connections between two manifolds. 
Informally, it shows that for any two points $p$ and $q$,  
a manifold $m(b)\in \calM^B$ that is close to both points enables a connection 
between two manifolds $m(a_p),m(a_q)\in \calM^A$ that 
are close to $p$ and $q$, respectively. 
}
{ 
\begin{proof}
$m(a)$ is~$\rho/\sqrt{2}$-intersecting for~$p$.
Hence, we know that the distance of~$a$ to~$p_A$ is less than~$\rho/\sqrt{2}$, 
the same holds for~$b$ and~$p_B$. 
Thus we can conclude (as demonstrated in Fig.~\ref{fig:ver_hor_intersection}): 
$$ 
dist(p,p')
=    \sqrt{dist(p_A,a)^2+dist(p_B,b)^2} 
\leq \rho.
$$
\end{proof}

The following lemma shows that for any two points $p$ and $q$,  
a manifold $m(b)\in \calM^B$ that is close to both points enables a connection 
between two manifolds $m(a_p),m(a_q)\in \calM^A$ that 
are close to $p$ and $q$, respectively. 
}

\begin{lemma}
\label{lem:local_path}
Let~$p,q \in C$ be two points such that~$dist(p,q) \leq \rho$
and let~$m(a_p), m(a_q)\in \calM^A$ be two~$\rho/\sqrt{2}$-intersecting
manifolds for~$p$ and~$q$ respectively. 
Let~$m(b)\in \calM^B$ be 
a manifold that is simultaneously~$\rho/\sqrt{2}$-intersecting
for~$p$ and~$q$ and let~$p'=(a_p,p_B)\in B_\rho^\calC(p)$ 
and~$q'=(a_q,q_B)\in B_\rho^\calC(q)$ be the projection 
of~$p$ and~$q$ on~$m(a_p)$ and~$m(a_q)$, respectively. 

There exists a path~$\gamma_{p', q'}$ between~$p'$ and~$q'$ 
such that~$Im(\gamma_{p', q'}) \subseteq 
(B_\rho^\calC(p) \cup B_\rho^\calC(q)) \cap (m(a_p)\cup m(b) \cup m(a_q))$, 
i.e., there is a path lying on the manifolds within the union of the balls. 
\end{lemma}

\begin{proof}
Let~$p''= m(a_p)\cap m(b) = (a_p,b)$ and~$q''= m(a_q)\cap m(b) = (a_q,b)$
denote the intersection point of~$m(a_p)$ and~$m(a_q)$ with~$m(b)$, 
respectively. Moreover, let~$p'''=(p_A,b)\in B_\rho^\calC(p)$ and 
$q'''=(q_A,b)\in B_\rho^\calC(q)$ denote the projections of~$p$ and~$q$ on~$m(b)$. 
We show that the path which is the concatenation of the segments 
$(p',p''),~(p'',p'''),~(p''',q'''),~(q''',q'')$ and~$(q'',q')$ 
lies on $(m(a_p)\cup m(b) \cup m(a_q))$ within the union of the balls $(B_\rho^\calC(p) \cup B_\rho^\calC(q))$. 
See Fig.~\ref{fig:path_example}. 

By Lemma~\ref{lem:intersection_point} the intersection points 
$p''$ and~$q''$ are inside~$B_\rho^\calC(p)$ and~$B_\rho^\calC(q)$, 
respectively. Thus, by convexity of each ball the segments 
$(p',p'') \subset m(q_p)$ and~$(q',q'') \subset m(a_q)$ as well as 
the segments~$(p'',p'''),(q'',q''') \subset m(b)$ are in  
$(B_\rho^\calC(p) \cup B_\rho^\calC(q))$.

It remains to show that~$(p''',q''') \subset m(b)$ is inside  
$(B_\rho^\calC(p) \cup B_\rho^\calC(q))$. 
Recall that~$dist(p,q) \leq \rho$ and therefore~$dist(p''',q''') \leq \rho$. 
Let~$\bar{p}$ be a point on the segment~$(p''',q''')$ 
that, w.l.o.g, is closer to~$p'''$. Thus~$dist(\bar{p},  p''') \leq \rho/2$.
The manifold~$m(b)$ is~$\rho/\sqrt{2}$-intersecting, thus~$dist(p, p''') \leq \rho/\sqrt{2}$.
As the segments~$(p,p''')$ and~$(p''',\bar{p})$ are perpendicular 
it holds: 
\begin{eqnarray*}
dist(p,\bar{p})  
&=   & \sqrt{dist(p,p''')^2 + dist(p''',\bar{p})^2}\\ 
&\leq& \sqrt{\rho^2/2 + \rho^2/4}\\
&<   &  \rho.
\end{eqnarray*}
\end{proof}

\begin{theorem}
\label{thm:completeness_3d}
Let~$p,q$ be points in $\Cfree$ such that there exists a collision-free path 
$\gamma_{p,q} \in \Gamma$ of length~$L$ and clearance~$\rho$ between 
$p$ and~$q$. 
Then the probability of the \MMS algorithm to return a path 
between~$p$ and~$q$ after 
generating~$n_A$ and~$n_B$ manifolds from families~$\calM^A$ and~$\calM^B$ as above, respectively is:
\begin{eqnarray*}
&     & Pr[(p,q) {\rm SUCCESS}]\\
&=    & 1 - Pr[(p,q) {\rm FAILURE}] \nonumber \\
&\geq & 
1 - \left\lceil \frac{L}{\rho} \right\rceil
\left[  
  \left( 1 -  \mu_A \right)^{n_A} 
  + 
  \left( 1 -  \eta_B \right)^{n_B}  
\right] \nonumber ,
\end{eqnarray*}
where~$\mu_A$ and~$\eta_B$ are some positive constants smaller than 1. 
\end{theorem}

\journal
{ 
The constants $\mu_A$ and~$\eta_B$ reflect the probability of a manifold
to be $\rho/\sqrt{2}$-intersecting for one or two nearby points, respectively.
The proof for Theorem~\ref{thm:completeness_3d} is rather technical. 
It involves using Lemma~\ref{lem:local_path} repeatedly for points along 
the path $\gamma_{p,q}$ of distance less than $\rho$. 
We omit the details and refer the reader to~\cite{SHH12-arxiv} for the  
full proof.
}
{
\begin{proof}
Let~$l = \left\lceil L/\rho\right\rceil$, there exists a 
sequence~$[p_0 \ldots p_\ell]$ such that~$p_i \in Im(\gamma_{p,q})$, 
$p_0=p$,~$p_\ell=q$,~$B_\rho(p_i)\in \Cfree$ and~$dist(p_i,p_{i+1}) \leq \rho$.
\MMS adds the manifolds~$m(p_A)$ and~$m(q_A)$ to the 
connectivity graph. 


Let~$A' \subset A$,~$|A'|=n_A$ and~$B' \subset B$,~$|B'|=n_B$, 
be the two point sets that define the manifolds~$M^{A'}$ and 
$M^{B'}$ that are used by the MMS algorithm. 
If there is a subset 
$\{ m(a_1) \ldots m(a_{\ell-1}) \} \subseteq M^{A'}$
and a subset  
$\{ m(b_1) \ldots m(b_{\ell-1}) \} \subseteq M^{B'}$
such that~$(p_i, p_{i+1}, m(a_i), m(b_i), m(a_{i+1}))$ fulfill the 
conditions of Lemma~\ref{lem:local_path} for~$i \in \{0 \ldots \ell-1\}$, 
then there exists a path from~$p$ to~$q$ in the \FSCs constructed by 
the MMS framework, namely the path which is the concatenation of paths 
constructed in Lemma~\ref{lem:local_path}. This implies that 
$p$ and~$q$ are in the same connected component of~$\calG$, which implies 
that MMS constructs a path in~$\Cfree$ from~$p$ to~$q$.

Let~$I_1 \ldots I_{\ell-1}$ be a set of indicator variables such that each 
$I_i$ witnesses the event that there is a~$\rho/\sqrt{2}$-intersecting 
manifold for~$p_i$ in~$M^{A'}$. 
(For $p_0$ and $p_\ell$ this is trivially the case due the explicit construction
of $m(p_A)$ and~$m(q_A)$.)
Let~$J_0 \ldots J_{\ell-1}$ be a set of indicator variables such that each~$J_i$ 
witnesses the event that there is a manifold in~$M^{B'}$ that is 
simultaneously~$\rho/\sqrt{2}$-intersecting for~$p_i$ and~$p_{i+1}$.
It follows that \MMS succeeds in answering the query 
$(p,q)$ if 
$I_i = 1$ for all~$ 1 \leq i \leq \ell-1$ and 
$J_j = 1$ for all~$ 0 \leq j \leq \ell-1$.
Therefore,
\begin{eqnarray}
 Pr[(p,q) {\rm FAILURE}] & \leq & 
     Pr \left( \vee^{\ell-1}_{i=1}(I_i = 0) 
                  \vee^{\ell-1}_{j=0}(J_j = 0) \right) \nonumber \\
               &\leq & 
      \sum^{\ell-1}_{i=1} Pr[I_i = 0] + \sum^{\ell-1}_{j=0} Pr[J_j = 0]. \nonumber 
\end{eqnarray}

\noindent
The events~$I_i = 0$ and~$J_j = 0$ are independent 
since the samples are taken independent. 
Thus the probability~$Pr[I_i = 0]$, i.e., that not even one of the~$n_A$
samples from~$A$ is~$\rho/\sqrt{2}$-intersecting for~$p_i$ is 
$(1-\mu_A)^{n_A}$, where~$\mu_A$ is the probability measure that a random
sample $a\in A$ defines a manifold that is $\rho/\sqrt{2}$-intersecting 
for a certain point $p\in\calC$. Thus, $\mu_A$ is obviously positive.   
Similarly,~$Pr[J_i = 0] = (1-\eta_B)^{n_B}$, 
where~$\eta_B$ is the probability measure that a random
sample $b\in B$ defines a manifold that is $\rho/\sqrt{2}$-intersecting 
for a two specific points $p,q\in\calC$ with $dist(p,q)<\rho$, that is,
it is proportional to the volume of the intersection 
$B^{B}_{\rho/\sqrt{2}}(p_B) \cap B^{B}_{\rho/\sqrt{2}}(q_B)$, which is positive
since the radius of the balls is larger than $\rho/2$. 
Since the sampling is uniform and independent: 
\begin{eqnarray}
 Pr[(p,q) {\rm FAILURE}] & \leq & 
 \left\lceil \frac{L}{\rho} - 1 \right\rceil  
  \left( 1 -  \mu_A\right)^{n_A} +
  \left\lceil \frac{L}{\rho} \right\rceil  
   \left( 1 -  \eta_B \right)^{n_B} \nonumber \\
              & \leq & 
  \left\lceil \frac{L}{\rho} \right\rceil
      \left[ 
        \left( 1 -  \mu_A \right)^{n_A} +
        \left( 1 -  \eta_B \right)^{n_B}  \right]. \nonumber 
\end{eqnarray}
\end{proof}

\noindent
It follows that as~$n_A$ and~$n_B$ tend to~$\infty$, the probability of 
failing to find a path under the conditions stated in 
Theorem~\ref{thm:completeness_3d} tends to zero.
}

\bigskip
\textbf{Recursive application }
The proof of Theorem~\ref{thm:completeness_3d} assumes that the samples are taken using  
full high-dimensional manifolds. However, Section~\ref{sec:two_robots} demonstrates a 
recursive application of \MMS where the approximate samples 
are generated by another application of \MMS.

In order to obtain a completeness proof for the two-level scheme let $\gamma$ be a path of clearance
$2\rho$. First, assume that the samples taken by the first level of MMS are exact. 
Applying Theorem~\ref{thm:completeness_3d} for $\gamma$ and 
$\rho$ shows that with sufficient probability \MMS would find a set $M'$ of manifolds
that would contain a path $\gamma'$. 
Since we required clearance $2\rho$ but relied on the tighter clearance $\rho$, 
it is guaranteed that $\gamma'$ still has clearance $\rho$. 
Now, each manifold $m'\in \calM'$ is actually only an approximation constructed by another 
application of \MMS. Thus, for each $m'\in \calM'$ apply Theorem~\ref{thm:completeness_3d}
to the subpath $\gamma'_{m'}=\gamma'\cap m'$ which has clearance $\rho$. 
Concatenation of all the resulting subpaths concludes the argument. 
Of course the parameters in the inequality in Theorem~\ref{thm:completeness_3d} change accordingly.

We remark that the recursive approach imposes a mild restriction on the sampling scheme 
as the sampling and the approximation must be somewhat coordinated. 
Since in theory $m(a)\cap m(b) = (a,b)$ we must ensure that points that we sample 
from $A$ are contained in every approximation of $m(b)\in \calM^B$ and vise versa. 
In our implementation this is ensured by restricting the set of possible angles
to those used to approximate $m(b)\in \calM^B$ (see Section~\ref{sec:two_robots}). 

\secspacea
\section{On the Dimension of Narrow Passages}
\secspaceb
\label{sec:discussion}

Consider the Pacman scenario illustrated in Fig.~\ref{fig:scenario_3}
of the experiments section. We obtain a narrow passage by increasing
the size of the square-shaped robot making it harder for the Pacman to 
swallow it. Fig.~\ref{fig:tightness} shows that our approach is significantly 
less sensitive to this tightening of the free space than the PRM algorithm. 
In order to explain this, let us take a closer 
look at the nature of the narrow passage for the tightest solvable case.

\begin{figure}
  \centering
  \includegraphics[scale=0.5]{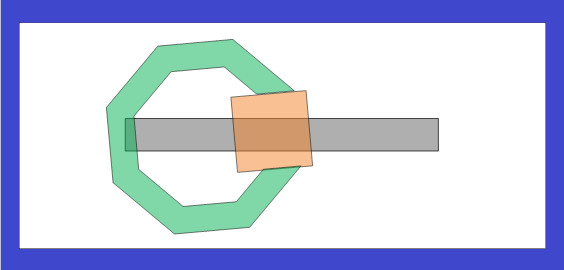}
  \caption{Tightest solvable Pacman scenario. 
    The gray rectangle shows valid placements of the 
    square such that the Pacman can engulf it without colliding 
    with the scene's bounding box.\vspace{-0.4cm}}
  \label{fig:narrowest}
\end{figure}

In order to get from the start placement to the goal placement, 
the Pacman must swallow the square, rotate around it and spit it out again. 
We concentrate on the swallowing motion. 
Fig.~\ref{fig:narrowest} depicts the tightest case, 
i.e., when the square robot fits exactly into the ``mouth'' of the Pacman. 
The gray rectangle indicates the positions 
of the reference point of the square such that there is a valid movement 
of the Pacman, considering the walls of the room, that will allow it to swallow the square robot 
(two-dimensional region, two parameters). 
The rotation angle of the square is also important (one additional parameter). 
The range of concurrently possible values for all three parameters is small 
but does not tend to zero either.
The passage becomes only narrow by the fact that the rotation angle 
of the Pacman must correlate exactly with the orientation of the square 
to allow for passing through the mouth. 
Moreover, the set of valid placements for the reference point of the Pacman while 
swallowing the square (other parameters being fixed) is a line, i.e., 
its~$x$ and~$y$ parameter values are coupled. 
Thus, the passage is a four-dimensional object as we have a tight coupling of 
two pairs of parameters in a six-dimensional \CS. 

The PRM approach has difficulties to sample in this passage 
since the measure tends to zero as the size of the square increases. 
On the other hand, for our approach the passage is only narrow 
with respect to the correlation of the two angles. 
%
As soon as the MMS samples an (approximated) volume that fixes the square robot such that the 
Pacman \emph{can} engulf it, the approximation of the volume just needs to include a 
horizontal slice of a suitable angle and the passage becomes 
evident in the corresponding Minkowski sum computation. 

\subsecspacea
\subsection{Definition of Narrow Passages}
\subsecspaceb
\label{subsec:narrow_passage_definition}

Intuition may suggest that narrow passages are tunnel-shaped. 
However, a one-dimensional tunnel in a high-dimensional \CSs would correspond to a simultaneous coupling of \emph{all} parameters, which is often not the case.
For instance, the discussion of the Pacman scenario shows that the passage is 
narrow but that it is still a four-dimensional volume, 
which proved to be a considerable advantage for our approach in the experiments.
Although some sampling based approaches try to take the dimension of a 
passage into account (see e.g.~\cite{DL11}) 
it seems that this aspect is not reflected by existing 
definitions that attempt to capture attributes of the \CS. 
Definitions such as $\epsilon$-goodness~\cite{KLMR98} and expansiveness~\cite{Hsu99} 
are able to measure the size of a narrow passage better than the  
clearance~\cite{Kavraki98} of a path, but neither incorporates the dimension
of a narrow passage in a very accessible way. 
Therefore, we would like to propose a new set of definitions that 
attempt to simultaneously grasp the narrowness and the dimension of a passage.

We start by defining the ``ordinary'' clearance of a path in \Cfree.
The characterization is based on the notion of homotopy classes of paths with respect to a set~$\Gamma_{s,t}$, i.e., 
the set of all paths starting at~$s$ and ending at~$t$. 
For a path~$\gamma_0\in\Gamma_{s,t}$ 
and its homotopy class~$\calH(\gamma_0)$ we define the clearance of the 
class as the largest clearance found among all paths in~$\calH(\gamma_0)$.

\begin{definition}
\label{def:clearance}
The \textbf{clearance} of a homotopy class~$\calH(\gamma_0)$ 
for~$\gamma_0 \in \Gamma_{s,t}$ is 
$$
\sup_{\gamma \in \calH(\gamma_0)}\\
\{~\sup \{~\rho>0~|~B_{\rho}^{d} \oplus Im(\gamma) \subseteq \Cfree~\}~\},
$$
\noindent where~$\oplus$ denotes the Minkowski sum of two sets, which is the vector sum of the sets. 
\end{definition}

By using a $d$-dimensional ball this definition treats all directions equally,
thus considering the passage of~$\calH(\gamma_0)$ to be a one-dimensional tunnel. 
We next refine this definition by using a $k$-dimensional 
disk, which may be placed in different orientations depending on the position along the path.

\begin{definition}
For some integer~$0 < \boldsymbol k \leq d$ the \textbf{$\boldsymbol k$-clearance} of~$\calH(\gamma_0)$ is:
\[
\sup_{\gamma \in \calH(\gamma_0)} 
\{
\rho>0|\forall t\in[0,1] \exists {\tt R}\in{\cal R}:\gamma(t)\oplus {\tt R}(t)B^{k}_{\rho} \subseteq \Cfree
\},
\]

\noindent where~${\cal R}$  is the set of $d$-dimensional rotation matrices 
and $B^{k}_{\rho}$ is the $k$-dimensional ball of radius $\rho$.
In case ${\tt R}$ is required to change continuously we talk about 
\textbf{continuous~$\boldsymbol k$-clearance}. 
\end{definition}
\noindent
Clearly, the~$k$-clearance of~$\calH(\gamma_0)$ for~$k = d$ is simply 
the clearance of~$\calH(\gamma_0)$. 
For decreasing values of~$k$, the~$k$-clearance of a homotopy class is a monotonically increasing sequence. 
We next define the dimension of a passage using this sequence, that is, 
we set the dimension to be the first~$k$ for which the clearance becomes 
{\em significantly larger}\footnote{We leave this notion informal as it might depend on the problem at hand.} than the original $d$-dimensional clearance.

\begin{definition}
A passage for~$\calH(\gamma_0)$ in $\R^d$ of clearance~$\rho$ 
(see Def.~\ref{def:clearance})
is called \textbf{$\boldsymbol {d-k+1}$-dimensional} if~$k$ is the largest index such that 
$k$-clearance$(\calH(\gamma_0)) \gg \rho$. 
If for every~$k$  $k$-clearance$(\calH(\gamma_0)) \not\gg \rho$ 
then we call the passage one-dimensional\footnote{For simplicity of definition we chose to stop at the largest index $k$ for which $k$-clearance $\gg  \rho$. 
One could contemplate alternative more elaborate definitions that keep on searching for even  larger clearance for smaller indices}. 
\end{definition}

\first{
\begin{figure}
  \centering
  \includegraphics[width=0.4\textwidth]{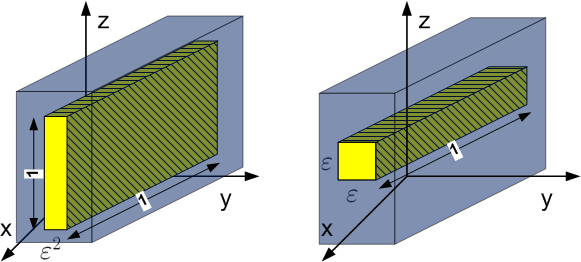}
  \caption{Two three-dimensional C-spaces consisting of a narrow passage (yellow) surrounded by obstacles (blue).}
  \label{fig:narrow_passages}
\end{figure}

For instance, both passages in Figure~\ref{fig:narrow_passages} have a measure of~$\varepsilon^2$ thus for a PRM like planner, 
sampling in either passage is equally hard as the probability of a uniform 
point sample to lie in either one of the narrow passages is proportional to~$\varepsilon^2$. 
However, the two passages are fundamentally different. 
The passage depicted on the right-hand side is a one-dimensional tunnel corresponding to a tight coupling of the three parameters. 
The passage depicted on the left-hand side is a two-dimensional flume which is much easier to intersect by a probabilistic approach that uses manifolds as samples. 
Our new definitions formally reveal this difference.
For $k$ equals 3, 2 and 1 the $k$-clearance of the right passage is
$\varepsilon$, $\sqrt{2}\varepsilon$ and  larger than $1$, respectively. For the left 
passage this sequence is $\varepsilon^2$ for $k=3$ and larger than $1$ for $k=2,1$ which characterizes 
the passage as two-dimensional. 
}{
\begin{figure}[t]
  \centering
   \begin{center}
       \includegraphics[width=0.6\textwidth]{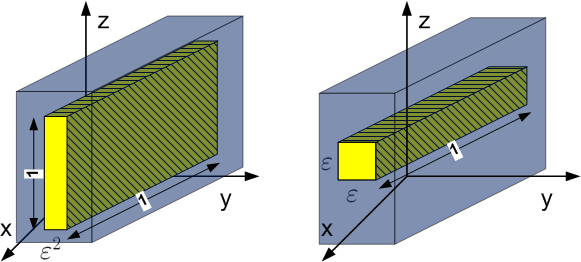}
   \end{center}
   \vspace{-3mm}
   \caption{\sf Narrow passages in three-dimensional configuration spaces: 
   							Both passages have a volume of $\varepsilon^2$ yet the 
   							left passage is two-dimensional while the 
   							right passage one-dimensional.}
   \label{fig:narrow_passages}
   \vspace{-6mm}
\end{figure}

Fig.~\ref{fig:narrow_passages} illustrates two three-dimensional \CSs consisting of a narrow passage (yellow) surrounded by obstacles (blue). 
Both passages have a measure of~$\varepsilon^2$ thus for a PRM like planner, 
sampling in either passage is equally hard as the probability of a uniform 
point sample to lie in either one of the narrow passages is proportional to~$\varepsilon^2$. 
However, the two passages are fundamentally different. 
The passage depicted on the right-hand side is a one-dimensional tunnel corresponding to 
a tight coupling of the three parameters. 
The passage depicted on the left-hand side is a two-dimensional flume which is much 
easier to intersect by a probabilistic approach that uses manifolds as samples. 
Our new definitions formally reveal this difference.
For $k$ equals 3, 2 and 1 the $k$-clearance of the right passage is
$\varepsilon$, $\sqrt{2}\varepsilon$ and  larger than $1$, respectively. For the left 
passage this sequence is $\varepsilon^2$ for $k=3$ and larger than $1$ for $k=2,1$ 
which characterizes the passage as two-dimensional. 
}
\subsecspacea
\subsection{Discussion}
\subsecspaceb

We believe that the definitions introduced in 
Section~\ref{subsec:narrow_passage_definition}, 
can be an essential component of a formal proof that shows the 
advantage of manifold samples over point samples in the presence of 
high-dimensional narrow passages. We sketch the argument briefly. 
Let \Cfree contain a narrow passage of dimension~$k$, that is, the passage has 
clearance $\rho$ and $k$-clearance $\lambda$, where $\lambda \gg \rho$. 
This implies that it is possible to place discs of dimension $k$ and 
radius $\lambda \gg \rho$ into the tight passage. The main argument is that 
for a random linear manifold of dimension $d-k$ the probability to 
hit such a disc is proportional to $\lambda$, which is much larger than $\rho$. 
The probability also depends on the
angle between the linear subspace containing the disc and the linear manifold. 
However, by choosing a proper set of manifold families it is possible to
guarantee the existence of at least one family for which this angle is bounded, 
independent of the orientation of the disk.

\secspacea
\section{Further Work}
\secspaceb
\label{sec:future_work}

The extension of MMS~\cite{SHRH11}  presented here is part of our 
on-going efforts towards the goal of creating a general scheme for exploring 
high-dimensional \CSs that is less sensitive to narrow passages than currently
available tools. 
As discussed in Section~\ref{subsec:preliminary_discussion} the original 
scheme imposes a set of conditions that in combination restrict an application
of MMS to rather low dimensions. In this paper we chose to relax condition 
{\bf C3}, for example by computing only approximations of three-dimensional manifolds. 
An alternative path is to relax condition {\bf C2}, for example by not sampling the 
manifolds uniformly and independently at random. 
This would enable the use of manifolds of low dimension 
as it allows to enforce intersection. 
Following this path we envision a single-query planner that explores a \CS 
in an RRT-like fashion. 
Using these extensions 
we wish to apply the scheme to a variety of difficult problems including assembly maintainability (part removal for maintenance~\cite{ZHKM08}) by employing a single-query variant of the scheme.

Another possibility is to explore other ways to compute approximative manifold samples,
for instance, the (so far) exact representations of \FSCs could be replaced by 
much simpler (and thus faster) but conservative\footnote{Approximated \FSCs are contained in \Cfree.} approximations. 
This is certainly applicable to manifold samples of dimension one or two
and should also enable manifold samples of higher dimensions. 
We remark that the use of approximations should not harm the probabilistic completeness 
as long as it is possible to refine the approximations such that they converge to the exact results.

In order to demonstrate the potential of the scheme, we adapted our motion planner to the problem of \emph{three-handed translational} assembly planning.
In assembly planning~\cite{WL94, HLW00}, we are given a collection of parts, and the goal is to assemble the parts into one (given) object. Typically, the problem is tackled by starting at the end configuration and recursively separating the object into sub-groups. 
Informally, the number of groups that may be considered \emph{simultaneously} is the number of hands used. 
The problem, which is known in general to be computationally hard (see, e.g.,~\cite{KK95}, 
has been studied extensively for two hands but little has been done for more. 
	
\begin{figure}
  \centering
  \subfloat
  [\sf ]
  {\label{fig:assembly1}
    \includegraphics[height=25mm]{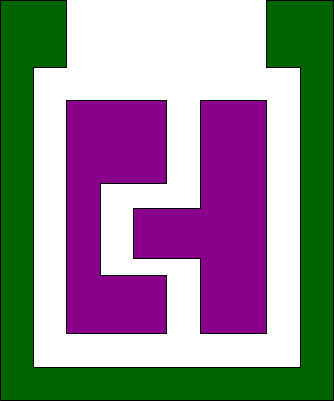}}
   \hfill
  \subfloat
  [\sf ]
  {\label{fig:assembly2}    
    \includegraphics[height=25mm]{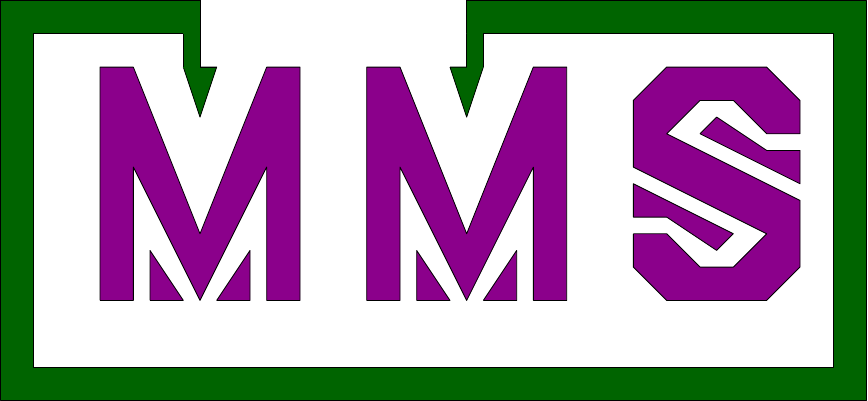}}
  \caption{\sf Three-handed assembly-planning scenarios. Static obstacle (first hand) is depicted in green and moving parts (second and third hands) are colored. In (b), at each iteration two moving parts are chosen randomly while the remaining parts are considered part of the static obstacles (if they were not disassembled in previous iterations).}
  \label{fig:assembly}
\end{figure}

The first assembly-planning problem we consider, depicted in Fig.~\ref{fig:assembly1} demonstrates a scenario where the two purple parts need to move in alternations in order to exit a surrounding part (the obstacle) in order to reach a disassembled configuration. This problem was solve by our planner within 37 seconds (average over 10 runs) and could not be solved by the PRM algorithm (which was terminated after 10 minutes). The RRT algorithm managed solving this scenario within 160 seconds  with a success rate of 50\% (if the solution was not found within 10 minutes the run was consiedered unsuccesfull; average over 10 runs).
The second assembly-planning problem, depicted in Fig.~\ref{fig:assembly2} demonstrates a scenario where multiple purple parts need
to be moved out of the surrounding green part. At each iteration, two purple parts are chosen at random and the planner attempts to translate both parts (as independent parts) out of the obstacle. The most interesting case occurs when the lower triangles have been removed and the two M-shaped parts need to translate out of the obstacle. In order for this to occur, the left M-shaped part needs to translate to the bottom-left corner  for the right M-shaped part to be able to translate out of the obstacle.
Our planner manages to plan this in under one second (average over 10 runs) while this case could not be solved by either the RRT or PRM algorithms (which were terminated after 10 minutes).
We note that this is not the traditional assembly-planning formulation as the parts are not touching each other. In order for a sampling-based algorithm (such as \MMS) to be applicable, some slack is required between the parts.
However, the slack can be much smaller when using \MMS as opposed to standard sampling-based planners.

%
%

Finally, we intend to extend 
the scheme to and experiment with motion-planning problems for highly-redundant robots as well as for fleets of robots, exploiting the
symmetries in the respective \CS.

For supplementary material, omitted here for lack of space, the reader is referred to our project web-page
 \url{http://acg.cs.tau.ac.il/projects/mms}.

\ifCLASSOPTIONcaptionsoff
  \newpage
\fi



%


\subsecspacea
\bibliography{bibliography}
\bibliographystyle{abbrv} 
\subsecspaceb

%


\begin{IEEEbiography}[{\includegraphics[width=1in,height=1.25in,clip,keepaspectratio]{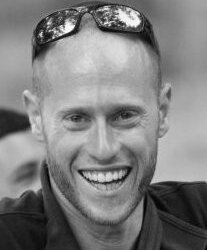}}]
{Oren Salzman} is a PhD-student at 
the School for Computer Science, Tel-Aviv University, Tel Aviv 69978, ISRAEL.
\email{orenzalz@post.tau.ac.il}
\end{IEEEbiography}

\begin{IEEEbiography}[{\includegraphics[width=1in,height=1.25in,clip,keepaspectratio]{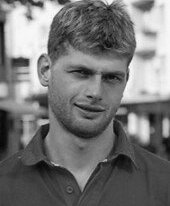}}]
{Michael Hemmer} 
was a post-doctoral fellow at 
the School for Computer Science, 
Tel-Aviv University, Tel Aviv 69978, ISRAEL during the time of this study 
and is now a researcher at the 
Institute of Operating Systems and Computer Networks, 
University of Technology Braunschweig, Braunschweig, Germany. 
\email{mhsaar@googlemail.com}
\end{IEEEbiography}

\begin{IEEEbiography}[{\includegraphics[width=1in,height=1.25in,clip,keepaspectratio]{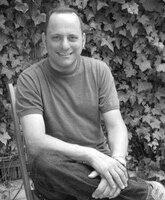}}]
{Dan Halperin} is a Professor at 
the School for Computer Science, 
Tel-Aviv University, Tel Aviv 69978, ISRAEL.
\email{danha@post.tau.ac.il}
\end{IEEEbiography}




\journal{
}{
\pagebreak
\appendix

\section*{ Critical Values for Rotating Robot}
\label{chap:critical_points}

We consider a polygonal robot rotating about a fixed reference point 
amidst polygonal obstacles. 
As the position of the robot is fixed the considered manifold is a 
vertical line in the configuration space, whose endpoints are identified. 
Thus, we parameterize the manifold with $\tau\in\PR$. 
On this manifold we are interested in the \FSCs, 
which are bounded by critical values. 
A critical value indicates a potential transition between \Cfree and \Cforb,
i.e., a configuration where the robot is in contact with an obstacles. 
More precisely: Either a robot's edge is in contact with
an obstacle's vertex or a robot's vertex is in contact with an obstacle's edge. 
These cases will be referred to as \emph{vertex-edge} contacts 
and \emph{edge-vertex} contacts, respectively. 
The rest of this section introduces the necessary notions to analyze the problem. 

A robot $R$ is a simple polygon with vertices $\{v_1,\dots,v_n\}$, where 
$v_i = (x_i,y_i)^T$ and edges $\{(v_1,v_2),  \dots (v_n,v_1)\}$.
We assume that the reference point of $R$ is located at the origin. 
The position of $R$ in the workspace is defined by a configuration 
$q = (r_q,\theta_q)$,  where $r_q = (x_q,y_q)^T$.  
Thus, $q$ maps the position of a vertex $v_i$ as follows: 
\[ v_i(q) = \left[ 
  \begin{array}{cc} 
    \cos\theta & -\sin\theta \\ 
    \sin\theta & \cos\theta
  \end{array} 
\right] v_i+r_q.\] 

\noindent
Given a fixed point $p = (x_p, y_p)^T$ 
we define in Equation~\ref{eq:par_T_1} the parameterization $(p, \tau) \in \R^2 \times \R\mathbb{P}^1$ which fixes the robot's reference point to a specific location.

\begin{equation}
  x_q = x_p, \ \ \ y_q = y_p, \ \ \   \theta_q = 2 \arctan \tau.
  \label{eq:par_T_1}
\end{equation}

\noindent
A parameterized vertex is represented in Equation (\ref{eq:v(q)}) 
\begin{equation}
  v_i(p,\tau) = 
  \frac{1}{1+\tau^2} 
  \left[ 
    \begin{array}{cc} 
      1-\tau^2 		& -2\tau		\\ 
      2\tau				& 1-\tau^2
    \end{array} 
  \right] 
  v_i + p.
  \label{eq:v(q)}
\end{equation}

\subsecspacea
\subsection*{Robot's vertex - Obstacle's edge}
\subsecspaceb

Let $r_q = (x_q, y_q)^T$ be the robot's fixed location.
Let $v_i$ be a robot's vertex and let 
$v_{o_1}$ and $v_{o_1}$ be the obstacle's edge's endpoints. 
The obstacle's edge can be parameterized as 
$e_o(s) = v_{o1} + s(v_{o2}-v_{o1})$, where $s \in [0,1]$.
If the distance between the fixed reference point and the obstacle's 
edge is larger than the distance between the robot's vertex and its
reference point, then the edge cannot impose a constraint. 
Namely all the edges $e$ such that $d(r_q,e) \geq d(r_q,v_i)$ may be 
filtered out.
A criticality occurs when the robot's vertex coincides with the edge $e$, 
thus $e(s) = v_i(r_q, \tau)$. 
This yields the following equalities:
\[
x_{o1} + s( x_{o2} - x_{o1} ) 
= 
\frac{1-\tau^2}{1+\tau^2} x_i - \frac{2 \tau y_i}{1+\tau^2} + x_q
\]
\[		
y_{o1} + s( y_{o2} - y_{o1} ) 
= 
\frac{2 \tau}{1+\tau^2} x_i + \frac{1-\tau^2}{1+\tau^2}y_i + y_q
\]

\noindent
Multiplication with $(1+\tau^2)$ yields
\[ 
(1+\tau^2)(x_{o1} + s( x_{o2} - x_{o1} ))
=
(1-\tau^2) x_i - (2 \tau) y_i + x_q(1+\tau^2)
\]
\[ 
(1+\tau^2)(y_{o1} + s( y_{o2} - y_{o1} ))
=
(2 \tau) x_i +  (1-\tau^2)y_i + y_q(1+\tau^2)
\]
\noindent
Or, 
\begin{eqnarray*}
& &
s(1+\tau^2)(x_{o2} - x_{o1})(y_{o2} - y_{o1})\\
&=&
[(1-\tau^2) x_i - (2 \tau) y_i + (1+\tau^2)(x_q - x_{o1})](y_{o2} - y_{o1})\\
\\
& &
s(1+\tau^2)(x_{o2} - x_{o1})(y_{o2} - y_{o1})\\
&=&
[(2 \tau) x_i +  (1-\tau^2)y_i + (1+\tau^2)(y_q - y_{o1})](x_{o2} - x_{o1})
\end{eqnarray*}

\noindent
denoting $\Delta_{ox} = x_{o2} - x_{o1}$, and $\Delta_{oy} = y_{o2} - y_{o1}$:
\begin{eqnarray*}
& &
[(1-\tau^2) x_i - (2 \tau) y_i + (1+\tau^2)(x_q - x_{o1})]\Delta_{oy}\\
&=&
[(2 \tau) x_i +  (1-\tau^2)y_i + (1+\tau^2)(y_q - y_{o1})]\Delta_{ox}
\end{eqnarray*}

%

\noindent
Finally:
\begin{equation}
  k_2 \tau^2 + k_1 \tau + k_0 = 0,
  \label{eq:v_e_1}
\end{equation}

\noindent
where
\begin{eqnarray*}
k_2 &=& (x_q - x_{o1} - x_i)\Delta_{oy} - (y_q - y_{o1} - y_i)\Delta_{ox} \\
k_1 &=& -2(y_i\Delta_{oy} +  x_i\Delta_{ox})	\\
k_0 &=& (x_q - x_{o1} + x_i)\Delta_{oy} - (y_q - y_{o1} +  y_i )\Delta_{ox} \\
\Delta_{ox} &=& x_{o2} - x_{o1}, \\
\Delta_{oy} &=& y_{o2} - y_{o1}.  \\
\end{eqnarray*}

\noindent
The solutions to Equation~\ref{eq:v_e_1} are two parameterized 
angles $\tau_j$, where $j\in\{1,2\}$. The corresponding 
position on the line can be identified by substituting 
$\tau$ with $\tau_j$ in one of the two equations for $s$,
where at least one is well defined as at least 
$(x_{o2} - x_{o1})$ or $(y_{o2} - y_{o1})$ does not vanish.

\[ 
s
=
\frac{(1-\tau^2) x_i - (2 \tau) y_i + (1+\tau^2)(x_q - x_{o1})}{ (x_{o2} - x_{o1})(1+\tau^2)}
\]
\[ 
s
=
\frac{(2 \tau) x_i +  (1-\tau^2)y_i + (1+\tau^2)(y_q - y_{o1})}{(y_{o2} - y_{o1})(1+\tau^2)}
\]

If $s_j \in [0,1]$, then the corresponding point is indeed on 
the edge and the value represents a potential transitions 
between \Cfree and \Cforb.


\subsecspacea
\subsection*{Robot's edge - Obstacle's vertex}
\subsecspaceb

Let $r_q = (x_q, y_q)^T$ be the fixed robot's location.
Let $v_1, v_2$ be the robot's vertex such that the robot's edge is defined 
as $e(s, r_q, \tau) = v_1(r_q, \tau) + s(v_2(r_q, \tau) - v_1(r_q, \tau))$ 
for $s \in [0,1]$ and $v_o$ be the obstacle's vertex. 
If the distance between the fixed reference point and the obstacle's vertex 
is larger than the distance between the robot's two vertices and its 
reference point, then the vertex cannot impose a constraint. 
Namely, all the obstacle vertices $v_o$ such that $d(r_q,v_0) \geq d(r_q,v_i)$ 
may be filtered out.
A criticality occurs when a point on the robot's edge coincides with the 
obstacle's vertex, namely for some $s \in [0,1]$:
\begin{eqnarray*}
v_{ox} &=& v_{1x} (p,\tau) + s(v_{2x} (p,\tau) - v_{1x} (p,\tau))\\
v_{oy} &=& v_{1y} (p,\tau) + s(v_{2y} (p,\tau) - v_{1y} (p,\tau))
\end{eqnarray*}

\noindent
Eliminating $s$ we obtain 
\begin{eqnarray*}
& & (v_{ox} - v_{1x} (p,\tau)) (v_{2y} (p,\tau) - v_{1y} (p,\tau))\\ 
&=& (v_{oy} - v_{1y} (p,\tau)) (v_{2x} (p,\tau) - v_{1x} (p,\tau)).
\end{eqnarray*}

\noindent
Denoting $\Delta_x = v_{2x} - v_{1x}$ and $\Delta_y = v_{2y} - v_{1y}$:
\begin{eqnarray*}
&&
(v_{ox} - v_{1x} (p,\tau)) 
(2\tau\Delta_x + (1-\tau^2)\Delta_y)\\
&=& 
(v_{oy} - v_{1y} (p,\tau)) 
((1-\tau^2)\Delta_x - 2\tau\Delta_y).
\end{eqnarray*}

\noindent
Denoting $a=(v_{ox} - x_q)$ and $b=(v_{oy} - y_q)$\\ and 
$k=v_{1x}v_{2y} - v_{2x}v_{1y}$, we obtain 
%
%
\begin{equation}
  l_2 \tau^2 + l_1 \tau + l_0 = 0,
  \label{eq:e_v_1}
\end{equation}

\noindent
where
\begin{eqnarray*}
  l_2 &=&   a \Delta_y - b \Delta_x + k, \\
  l_1 &=&-2(a \Delta_x + b \Delta_y),	\\
  l_0 &=& -l_2 +2k, \\
  \Delta_x &=& v_{2x} - v_{1x}, \\
  \Delta_y &=& v_{2y} - v_{1y}.  \\
\end{eqnarray*}

Similarly to the case of a robot's vertex in contact with an obstacles edges
   the solutions to Equation~\ref{eq:e_v_1} are two parameterized angles 
$\tau_j$, for $j\in\{1,2\}$, that represent potential transitions between \Cfree and \Cforb.
Again, these angles may represent intersections that are not on the robot's edge 
but on the line supporting the edge. Namely, if we obtain $s_j$ by 
plugging $\tau_j$ into either one of the two following equations and 
$s_j \notin [0,1]$ then we should not consider the corresponding~$\tau_j$.

\begin{eqnarray*}
s&=& \frac{v_{ox} -v_{1x} (p,\tau)}{v_{2x} (p,\tau) - v_{1x} (p,\tau)}\\
s&=& \frac{v_{oy} -v_{1y} (p,\tau)}{v_{2y} (p,\tau) - v_{1y} (p,\tau)}
\end{eqnarray*}

}

\end{document}